\documentclass{svproc}
\usepackage{url}
\usepackage{pgfplots}
\pgfplotsset{compat=1.3} 
\usepackage{tikz}
\usepackage{subcaption}
\usepackage{booktabs}
\usepackage{multirow}
\usepackage{chngpage}
\usepackage{tabularx}

\let\llncssubparagraph\subparagraph
\let\subparagraph\paragraph
\usepackage[compact]{titlesec}

\titlespacing*{\section}
  {0pt}{0.55\baselineskip}{0.55\baselineskip}
 \titlespacing*{\subsection}
  {0pt}{0.55\baselineskip}{0.55\baselineskip}
  \titlespacing*{\paragraph}
  {0pt}{0.25\baselineskip}{0.25\baselineskip}
\let\subparagraph\llncssubparagraph

\begin{document}
\mainmatter              
\title{Energy Consumption Analysis of pruned Semantic Segmentation Networks on an Embedded GPU}
\titlerunning{Pruned Networks on an Embedded GPU}  
%
\author{Hugo Tessier\inst{1}\inst{2} \and Vincent Gripon\inst{2}
Mathieu Léonardon\inst{2} \and Matthieu Arzel\inst{2} \and \\ David Bertrand\inst{1} \and Thomas Hannagan\inst{1} }
\authorrunning{Hugo Tessier et al.} 
%
\tocauthor{Hugo Tessier, Vincent Gripon, Mathieu Léonardon, Matthieu Arzel,
David Bertrand, Thomas Hannagan}
\institute{Stellantis, Centre Technique Vélizy, Vélizy-Villacoublay 78140, France\\
\email{firstname.lastname@stellantis.com},
\and
IMT Atlantique, Lab-STICC, UMR CNRS 6285, F-29238, France \\ \email{firstname.lastname@imt-atlantique.fr}}

\maketitle              

\begin{abstract}
Deep neural networks are the state of the art in many computer vision tasks. Their deployment in the context of autonomous vehicles is of particular interest, since their limitations in terms of energy consumption prohibit the use of very large networks, that typically reach the best performance. A common method to reduce the complexity of these architectures, without sacrificing accuracy, is to rely on pruning, in which the least important
portions are eliminated. There is a large literature on the subject, but interestingly few works have measured the actual impact of pruning on energy. In this work, we are interested in measuring it in the specific context of semantic segmentation for autonomous driving, using the Cityscapes dataset. To this end, we analyze the impact of recently proposed structured pruning methods when trained architectures are deployed on a Jetson Xavier embedded~GPU.
\keywords{Deep Learning, Compression, Pruning, Hardware Implementations, Energy}
\end{abstract}
\section{Introduction}

In the course of the last decade, deep neural networks have become a staple in many computer vision tasks such as classification, detection or semantic segmentation. This makes them especially interesting for the field of autonomous vehicles. Cityscapes~\cite{cordts2016cityscapes} is a semantic segmentation dataset designed for this field. It presents a significant challenge due to its high resolution, its size and the variety of images and classes.

For this dataset, state of the art is held by networks such as HRNets~\cite{sun2019high}, which are composed of a large number of parameters and operations during inference and produce large intermediate products that cause a large memory footprint. Such costs are prohibitive for embedded devices and thus compromise their usage in autonomous vehicles.

To reduce the cost of a network, the domain of neural networks compression presents multiple different methods, such as quantization~\cite{courbariaux2015binaryconnect}, distillation~\cite{hinton2015distilling} or pruning~\cite{han2015deep}. In this article we focus on pruning and more especially on ``structured'' filter pruning~\cite{liu2017learning} that removes entire neurons in convolution layers of deep neural networks.
Structured pruning can be a challenge to implement when applying different pruning rates for each layer of a network that includes long-range dependencies between layers because of, for example, residual connections~\cite{he2016deep}. Therefore, slimming down networks is a distinct issue, while most works in the field stick to putting weights to zero to measure, instead, the impact of pruning on a network's performance.

Our article aims at providing a solution to leverage irregular structured sparsity in HRNet-48 and measure the gains in energy consumption on a NVIDIA Jetson AGX Xavier embedded GPU. We report results using two different pruning methods of the literature and show how each of them allows for a different trade-off between performance and cost for the same network architecture. We also compare these results to that of non-pruned HRNet-32 and HRNet-18, and discuss the implications of these results in the general discussion on the ability of pruning to produce efficient architectures.

\section{Related Works}

\subsection{Semantic Segmentation Networks}

At first, deep semantic segmentation neural networks were built by concatenating a classification network and a decoder. The convolution layers of the (usually pretrained) encoder classification network produce intermediate representations that are successively fed to the decoder that upsamples and sums them together.

This principle was introduced by Long et al.~\cite{long2015fully} (FCN) and then perfected by Ronneberger et al.~\cite{ronneberger2015u} (U-net) that cemented the symmetry between encoder and decoder that can be found in other networks, such as SegNet~\cite{badrinarayanan2017segnet}.
Dilated (or ``atrous'') convolutions~\cite{yu2015multi} led to another family of semantic segmentation networks that, instead of upsampling and summing together intermediate representations of decreasing resolution, managed to alter a classification network so that said resolutions remains high. Such principle can be found among DeepLab networks~\cite{chen2017deeplab}, that also introduce the principle of pyramid pooling (i.e., operating on multiple resolutions at once to extract information at various scales). 

Both ideas of pyramid pooling and maintaining high resolution gave birth to HRNet~\cite{sun2019high}, which, instead of being a modified classification network, is fully designed for such a task. Variants of it, with an additional decoder, are state of the art~\cite{yuan2020object}. However, HRNet tends to be a very complex network, whose cost is unsuitable for embedded hardware. Some works draw inspiration from HRNet to design networks more adapted to such hardware~\cite{hong2021deep}. Our approach is to study how pruning could reduce the cost of HRNet.

\subsection{Neural Network Pruning}

First introduced during the late 80's~\cite{lecun1989optimal}, pruning is now a widespread method to reduce the cost of deep neural networks~\cite{han2015learning}. Since removing weights in an unstructured manner leads to sparse tensors that are difficult to optimize efficiently~\cite{liu2021s2ta,ma2021non}, it is possible instead to remove whole neurons, i.e. filters in the case of convolution layers~\cite{liu2017learning}. Indeed, such kind of removal alters input and output dimensions of layers and it is necessary to remove inter-dependant weights between layers in a consistent way.

This is rather straightforward to do in simple networks such as VGG~\cite{simonyan2014very}, but networks such as HRNets present much more intricate dependencies between layers, partly because of residual connections~\cite{he2016deep}. Indeed, such connections imply a certain regularity in the dimensions of intermediate products of the network. In the case of global pruning, where each layer can be pruned at different rates to get a global count of parameters (in opposition to local pruning that prunes equally every layer), pruning can break this regularity, which means that summing these tensors is not possible without scattering them into tensors of the same dimension. Multiple works solve this problem using different kind of operators~\cite{he2017channel}.

\section{Method}

Our work focuses on pruning HRNet. Indeed, both its significance in the state of the art and its prohibitive cost in memory and computation power make it an especially relevant subject for experimentation to measure how pruning can make semantic segmentation suitable for embedded hardware.

We used the Cityscapes dataset~\cite{cordts2016cityscapes} to train our networks. Indeed, Cityscapes is a popular dataset in the scope of autonomous vehicles and features high resolution images and annotations for 19 different classes (among others that are not used during training). All the results featured in this article concern networks we trained ourselves.

We chose two different structured pruning methods: Slimming~\cite{liu2017learning}, an influential method in the literature, and SWD~\cite{tessier2022rethinking}, that is more recent and has shown a certain ability to give competitive results on multiple networks or tasks with different kinds of pruning structures.

All our measurements were run on a NVIDIA Jetson AGX Xavier embedded GPU under the ``30W All'' mode, with JetPack SDK 5.0, CUDA 11.4.14, cuDNN 8.3.2, TensorRT 8.4.0 EA and running inference with ONNX Runtime 1.12.0 while using the TensorRT execution provider and using the tegrastats utility to measure power consumption. The number of MAC operations is counted using the ONNX Operations Counter tool.

\subsection{Training Conditions}

Our HRNet networks are trained during 200 epochs with a batch size of 10 on 3 NVIDIA Quadro K6000 GPUs, using the Pytorch 1.9.0 framework with CUDA 11.3. We use the SGD optimizer with weight decay set to $5\cdot10^{-4}$, momentum set to 0.9 and base learning rate set to 0.01. We decrease learning rate using the poly policy with the power of 2 (i.e. learning rate is reduced by $(1-\frac{current\_epoch}{epochs})^2$ at each epoch). We use the RMI loss~\cite{zhao2019region} and report accuracy in term of mean intersection over union (mIoU). Most of these hyperparameters come from the original papers of HRNet.

During testing, input images are of size $3\times1024\times2048$. During training, they are randomly cropped and resized, with a scale of $[0.5, 2]$, to $3\times512\times1024$. Data augmentation also involves random flips, random Gaussian blur, and color jittering. All samples from the dataset are normalized. More implementary precisions can be found consulting the source code available at \url{https://github.com/HugoTessier-lab/Neural-Network-Shrinking.git}.

Our baseline performance for HRNet-48 shows significant differences with the results announced in the original paper~\cite{sun2019high}. Indeed, using recent versions of Python and Pytorch (that we need for our tools to work properly), we were unable to reproduce the original results, even when applying the same training conditions or even when using available pretrained networks and the original source code. When using a pretrained HRNet-48 on Cityscapes, the best results we got were obtained with our implementation and lead to a mIoU accuracy of 73\%. Therefore, we conclude that the performance we get with our own training, which is of 77\% in mIoU, is good enough not to harm the conclusions of our article in any way.

\subsection{Pruning Strategies}\label{strategies}

\paragraph{\textbf{Pruning Structure and Target}:} We prune convolution filters and all the weights that depend on them. This includes corresponding batch normalization coefficients and kernels in filters of following layers. We also take into account long-range dependencies introduced by residual connections. The tensors of the network's weights are then reshaped so that the remaining parameters allocated in memory, the pruning target and the count of weights that still contribute to the network's function perfectly match.

\paragraph{\textbf{Pruning criterion}:} We prune filters on the basis of the magnitude of the learnt multiplicative coefficient in its corresponding batch-normalization layer~\cite{liu2017learning}.

\paragraph{\textbf{Pruning methods}}
\begin{itemize}
    \item \textbf{Slimming}: During training, batch-normalization layers are penalized by a smooth-$\mathcal{L}_1$ norm of their multiplicative coefficient, in order to enforce sparsity. Networks are pruned in three steps, with a linearly increasing pruning rate until the final one is reached. After each pruning step, the network is fine-tuned during 20 epochs. Such an iterative process helps preventing layer collapse~\cite{tanaka2020pruning}. Once the final pruning rate is reached, the network is re-trained following a warm-restart schedule, which can be called LR-Rewinding~\cite{renda2020comparing}.
    \item \textbf{SWD}: Since SWD prunes progressively throughout pruning, it does not need iterations. Instead we prune directly at the intended final pruning rate once the networks have been trained under the constraint of SWD's penalty. The hyperparameters $a_{min}$ and $a_{max}$ of SWD are set to $1\cdot10^{-1}$ and $1\cdot10^{10}$ respectively, because with weight decay set to $5\cdot10^{-4}$, learning rate to 0.01 and the aforementioned scheduler, we get $a_{min}\times WD\times LR = 0.1\times WD \times LR < WD\times LR$ and $a_{max}\times WD \times LR \times \prod_{n=1}^{200}\left(1-\frac{n}{200}\right)^2 \approx 1\cdot10^{10} \times 1.25\cdot10^{-10} \approx 1$, which means that SWD starts off as negligible and ends up applying a penalty to targeted weights that almost equals their magnitude, thus pruning them. After that, LR-Rewinding is applied.
\end{itemize}

\subsection{ONNX Implementation details}

The ONNX format has the clear advantage of being supported by a wide array of frameworks, that themselves can be ported on many different hardware. Therefore, gains observed in our experiments can be reproduced on other devices.

Structured pruning can introduce some dimensional discrepancies, which can be solved by introducing operators to reshape tensors coming from layers that were pruned differently~\cite{he2017channel}. In our implementation, we chose to insert a \textit{ScatterND} ONNX operator, that inserts the data from a tensor into another whose dimension matches that of the layers that will take this tensor as an input. This operator is placed before any tensor addition operation to prevent any incompatibility in dimensions.
This solves the problem but counts a drawback, that can be however put into perspective. Indeed, this operator needs additional temporary empty feature maps before summation, which introduces an additional cost in memory space and latency at inference. This problem is inevitable with our paradigm and can be found in previous works~\cite{he2017channel}. Preventing this problem would require rearranging the network in a much more complicated way, which we wished not to dwell into yet.

This implementation is the one that gave us the best performance in inference time, energy consumption and size of the network. However, it still has a non-negligible impact on performance, as can be seen in our experiments. This cost is likely due to the unconventional nature of this operator in neural networks, that makes it poorly supported by most frameworks. Indeed, TensorRT treats it as a ``foreign node'' that is accelerated using a library, Myelin, distinct from those used for all other nodes. The fact that the \textit{Scatter} operator and its variants are only supported since TensorRT 8.0 hints that the acceleration of these operations still has room for improvement. Yet, this overall configuration is the one that performed the best during our preliminary experiments.

\section{Experiments}

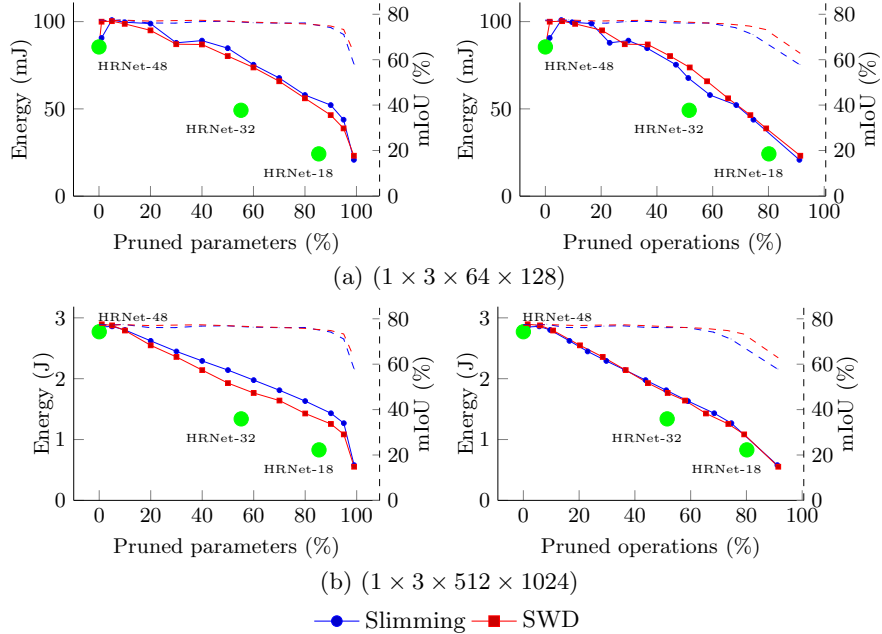
\begin{figure}[ht]
    \centering
    \vskip\baselineskip
    \begin{subfigure}{\linewidth}
    	\centering
    	\begin{tikzpicture}
    	\begin{scope}[scale=0.9]
    	\begin{axis}[
    	xlabel=Pruned parameters (\%),
    	ymin=0,
    	mark size = 1pt,
    	xtick={0,20,40,60,80,100},
    	ylabel=Energy (mJ),
    	ylabel shift = -6 pt,
    	width=0.5\textwidth,
    	height=1.75in,
    	legend entries={Slimming, SWD},
    	legend to name=named,
    	legend columns=-1,
    	legend style = {draw=none},
    	xtick pos=left,
    	ytick pos=left,
    	x axis line style={draw=none, insert path={(axis cs:\pgfkeysvalueof{/pgfplots/xmin},0,0) edge (axis cs:\pgfkeysvalueof{/pgfplots/xmax},0,0)}},
    	y axis line style={draw=none, insert path={(axis cs:\pgfkeysvalueof{/pgfplots/xmin},\pgfkeysvalueof{/pgfplots/ymin},0) edge (axis cs:\pgfkeysvalueof{/pgfplots/xmin},\pgfkeysvalueof{/pgfplots/ymax},0)}},
    	]
    	
    	\addplot coordinates
    	{
    		(0, 85.482)
    		(1, 90.632)
    		(5, 100.874)
    		(10, 99.632)
    		(20, 98.861)
    		(30, 87.846)
    		(40, 89.114)
    		(50, 84.769)
    		(60, 75.298)
    		(70, 67.685)
    		(80, 57.934)
    		(90, 52.175)
    		(95, 43.795)
    		(99, 20.813)
    	};
    	\addplot coordinates
    	{
    		(0, 85.482)
    		(1, 99.941)
    		(5, 100.256)
    		(10, 98.702)
    		(20, 94.979)
    		(30, 87.055)
    		(40, 86.898)
    		(50, 80.27)
    		(60, 73.735)
    		(70, 65.823)
    		(80, 56.063)
    		(90, 46.43)
    		(95, 38.825)
    		(99, 23.152)
    	};
    	
    	
    	\addplot[only marks,
    	mark=*,
    	mark size=3pt,
    	green,
    	nodes near coords = HRNet-32,
    	nodes near coords align={vertical},
    	point meta=y,
    	every node near coord/.append style={font=\tiny, yshift=-4.5mm, xshift=-3mm, color=black},
    	] coordinates
    	{
    		(55.14, 49.199)
    	};
    	
    	\addplot[
    	only marks,
    	mark=*,
    	mark size=3pt,
    	green,
    	nodes near coords = HRNet-18,
    	nodes near coords align={vertical},
    	point meta=y,
    	every node near coord/.append style={font=\tiny, yshift=-4.5mm, xshift=-3mm, color=black},
    	] coordinates
    	{
    		(85.36, 24.22)
    	};
    	
    	\addplot[
    	only marks,
    	mark=*,
    	mark size=3pt,
    	green,
    	nodes near coords = HRNet-48,
    	nodes near coords align={vertical},
    	point meta=y,
    	every node near coord/.append style={font=\tiny, yshift=-4.5mm, xshift=5mm, color=black},
    	] coordinates
    	{
    		(0, 85.482)
    	};
    	
    	\end{axis}
    	\begin{axis}[
    	ymin=0,
    	mark size = 0pt,
    	hide x axis,
    	axis y line*=right,
    	ylabel shift = -5 pt,
    	width=0.5\textwidth,
    	height=1.75in,
    	ylabel={mIoU (\%)},
    	ylabel near ticks,
    	dashed,
    	xtick pos=left,
    	ytick pos=right
    	]
    	
    	\addplot coordinates
    	{
    		(01, 77.4)
    		(05, 77.2)
    		(10, 77.56 ) 
    		(20, 76.17) 
    		(30, 76.10 ) 
    		(40, 76.78) 
    		(50, 76.92) 
    		(60, 76.29) 
    		(70, 76.10) 
    		(80, 76.17) 
    		(90, 74.15) 
    		(95 , 71.1)
    		(99 , 57.9)
    	};
    	\addplot coordinates
    	{
    		(1, 77.1)
    		(5, 77.1)
    		(10, 77.44)
    		(20, 77.02)
    		(30, 77.22)
    		(40, 77.35)
    		(50, 76.97)
    		(60, 76.53)
    		(70, 76.19)
    		(80, 75.77)
    		(90, 74.82)
    		(95, 73.3)
    		(99 , 62.8)
    	};
    	
    	\end{axis}
    	\end{scope}
    	\end{tikzpicture}
    	\begin{tikzpicture}
    	\begin{scope}[scale=0.9]
    	\begin{axis}[
    	xlabel=Pruned operations (\%),
    	ymin=0,
    	mark size = 1pt,
    	xtick={0,20,40,60,80,100},
    	ylabel=Energy (mJ),
    	ylabel shift = -6 pt,
    	width=0.5\textwidth,
    	height=1.75in,
    	xtick pos=left,
    	ytick pos=left,
    	x axis line style={draw=none, insert path={(axis cs:\pgfkeysvalueof{/pgfplots/xmin},0,0) edge (axis cs:\pgfkeysvalueof{/pgfplots/xmax},0,0)}},
    	y axis line style={draw=none, insert path={(axis cs:\pgfkeysvalueof{/pgfplots/xmin},\pgfkeysvalueof{/pgfplots/ymin},0) edge (axis cs:\pgfkeysvalueof{/pgfplots/xmin},\pgfkeysvalueof{/pgfplots/ymax},0)}},
    	]
    	
    	\addplot coordinates
    	{
    		
    		(0, 85.482)
    		(1.54, 90.632)
    		(5.7, 100.874)
    		(9.6, 99.632)
    		(16.52, 98.861)
    		(23.05, 87.846)
    		(29.76, 89.114)
    		(36.49, 84.769)
    		(46.83, 75.298)
    		(51.25, 67.685)
    		(59.03, 57.934)
    		(68.45, 52.175)
    		(74.61, 43.795)
    		(91.1, 20.813)
    		
    	};
    	\addplot coordinates
    	{
    		(0, 85.482)
    		(1.56, 99.941)
    		(5.98, 100.256)
    		(10.67, 98.702)
    		(20.31, 94.979)
    		(28.47, 87.055)
    		(36.71, 86.898)
    		(44.69, 80.27)
    		(51.77, 73.735)
    		(57.98, 65.823)
    		(65.5, 56.063)
    		(73.51, 46.43)
    		(79.19, 38.825)
    		(91.42, 23.152)
    	};
    	
    	
    	\addplot[only marks,
    	mark=*,
    	mark size=3pt,
    	green,
    	nodes near coords = HRNet-32,
    	nodes near coords align={vertical},
    	point meta=y,
    	every node near coord/.append style={font=\tiny, yshift=-4.5mm, xshift=-3mm, color=black},
    	] coordinates
    	{
    		(51.57, 49.199)
    	};
    	
    	\addplot[
    	only marks,
    	mark=*,
    	mark size=3pt,
    	green,
    	nodes near coords = HRNet-18,
    	nodes near coords align={vertical},
    	point meta=y,
    	every node near coord/.append style={font=\tiny, yshift=-4.5mm, xshift=-3mm, color=black},
    	] coordinates
    	{
    		(80.15, 24.22)
    	};
    	
    	\addplot[
    	only marks,
    	mark=*,
    	mark size=3pt,
    	green,
    	nodes near coords = HRNet-48,
    	nodes near coords align={vertical},
    	point meta=y,
    	every node near coord/.append style={font=\tiny, yshift=-4.5mm, xshift=5mm, color=black},
    	] coordinates
    	{
    		(0, 85.482)
    	};
    	
    	\end{axis}
    	\begin{axis}[
    	ymin=0,
    	mark size = 0pt,
    	hide x axis,
    	axis y line*=right,
    	ylabel shift = -5 pt,
    	width=0.5\textwidth,
    	height=1.75in,
    	ylabel={mIoU (\%)},
    	ylabel near ticks,
    	dashed,
    	xtick pos=left,
    	ytick pos=right
    	]
    	
    	\addplot coordinates
    	{
    		(1.54 , 77.4)
    		(5.7 , 77.2)
    		(9.6, 77.56 ) 
    		(16.52, 76.17) 
    		(23.05, 76.10 ) 
    		(29.76, 76.78) 
    		(36.49, 76.92) 
    		(43.83, 76.29) 
    		(51.25, 76.10) 
    		(59.03, 76.17) 
    		(68.45, 74.15) 
    		(74.61 , 71.1)
    		(91.1 , 57.9)
    	};
    	\addplot coordinates
    	{
    		(1.56, 77.1)
    		(5.98, 77.1)
    		(10.67, 77.44)
    		(20.31, 77.02)
    		(28.47, 77.22)
    		(36.71, 77.35)
    		(44.69, 76.97)
    		(51.77, 76.53)
    		(57.98, 76.19)
    		(65.5, 75.77)
    		(73.51, 74.82)
    		(79.19 , 73.3)
    		(91.42 , 62.8)
    	};
    	
    	\end{axis}
    	\end{scope}
    	\end{tikzpicture}
    	\vspace{-5pt}
    	\caption{$(1\times 3\times 64\times 128)$}
    	\vspace{-8pt}
    \end{subfigure}%
    \vskip\baselineskip

    \begin{subfigure}{\linewidth}
    	\centering
    	\begin{tikzpicture}
    	\begin{scope}[scale=0.9]
    	\begin{axis}[
    	xlabel=Pruned parameters (\%),
    	ymin=0,
    	mark size = 1pt,
    	xtick={0,20,40,60,80,100},
    	ylabel=Energy (J),
    	ylabel shift = -6 pt,
    	width=0.5\textwidth,
    	height=1.75in,
    	xtick pos=left,
    	ytick pos=left,
    	x axis line style={draw=none, insert path={(axis cs:\pgfkeysvalueof{/pgfplots/xmin},0,0) edge (axis cs:\pgfkeysvalueof{/pgfplots/xmax},0,0)}},
    	y axis line style={draw=none, insert path={(axis cs:\pgfkeysvalueof{/pgfplots/xmin},\pgfkeysvalueof{/pgfplots/ymin},0) edge (axis cs:\pgfkeysvalueof{/pgfplots/xmin},\pgfkeysvalueof{/pgfplots/ymax},0)}},
    	]
    	
    	\addplot coordinates
    	{
    		(0, 2.771487)
    		(1, 2.849576)
    		(5, 2.85994)
    		(10, 2.800701)
    		(20, 2.621482)
    		(30, 2.447645)
    		(40, 2.291658)
    		(50, 2.141073)
    		(60, 1.976244)
    		(70, 1.809942)
    		(80, 1.63196)
    		(90, 1.431767)
    		(95, 1.268395)
    		(99, 0.579087)
    	};
    	\addplot coordinates
    	{
    		(0, 2.771487)
    		(1, 2.894104)
    		(5, 2.876006)
    		(10, 2.791686)
    		(20, 2.548079)
    		(30, 2.357087)
    		(40, 2.14092)
    		(50, 1.927579)
    		(60, 1.76483)
    		(70, 1.640623)
    		(80, 1.428109)
    		(90, 1.255498)
    		(95, 1.084564)
    		(99, 0.552862)
    	};
    	
    	\node[anchor=south ] at (axis cs: \pgfkeysvalueof{/pgfplots/xmin},-0.33*\pgfkeysvalueof{/pgfplots/ymax}) {(\textbf{c})};
    	
    	\addplot[only marks,
    	mark=*,
    	mark size=3pt,
    	green,
    	nodes near coords = HRNet-32,
    	nodes near coords align={vertical},
    	point meta=y,
    	every node near coord/.append style={font=\tiny, yshift=-4.5mm, xshift=-3mm, color=black},
    	] coordinates
    	{
    		(55.14, 1.339235)
    	};
    	
    	\addplot[
    	only marks,
    	mark=*,
    	mark size=3pt,
    	green,
    	nodes near coords = HRNet-18,
    	nodes near coords align={vertical},
    	point meta=y,
    	every node near coord/.append style={font=\tiny, yshift=-4.5mm, xshift=-3mm, color=black},
    	] coordinates
    	{
    		(85.36 , 0.829931)
    	};
    	
    	\addplot[
    	only marks,
    	mark=*,
    	mark size=3pt,
    	green,
    	nodes near coords = HRNet-48,
    	nodes near coords align={vertical},
    	point meta=y,
    	every node near coord/.append style={font=\tiny, yshift=0.5mm, xshift=5mm, color=black},
    	] coordinates
    	{
    		(0, 2.771487)
    	};
    	
    	\end{axis}
    	\begin{axis}[
    	ymin=0,
    	mark size = 0pt,
    	hide x axis,
    	axis y line*=right,
    	ylabel shift = -5 pt,
    	width=0.5\textwidth,
    	height=1.75in,
    	ylabel={mIoU (\%)},
    	ylabel near ticks,
    	dashed,
    	xtick pos=left,
    	ytick pos=right
    	]
    	
    	\addplot coordinates
    	{
    		(01, 77.4)
    		(05, 77.2)
    		(10, 77.56 ) 
    		(20, 76.17) 
    		(30, 76.10 ) 
    		(40, 76.78) 
    		(50, 76.92) 
    		(60, 76.29) 
    		(70, 76.10) 
    		(80, 76.17) 
    		(90, 74.15) 
    		(95 , 71.1)
    		(99 , 57.9)
    	};
    	\addplot coordinates
    	{
    		(1, 77.1)
    		(5, 77.1)
    		(10, 77.44)
    		(20, 77.02)
    		(30, 77.22)
    		(40, 77.35)
    		(50, 76.97)
    		(60, 76.53)
    		(70, 76.19)
    		(80, 75.77)
    		(90, 74.82)
    		(95, 73.3)
    		(99 , 62.8)
    	};
    	
    	\end{axis}
    	\end{scope}
    	\end{tikzpicture}
    	\begin{tikzpicture}
    	\begin{scope}[scale=0.9]
    	\begin{axis}[
    	xlabel=Pruned operations (\%),
    	ymin=0,
    	mark size = 1pt,
    	xtick={0,20,40,60,80,100},
    	ylabel=Energy (J),
    	ylabel shift = -6 pt,
    	width=0.5\textwidth,
    	height=1.75in,
    	xtick pos=left,
    	ytick pos=left,
    	x axis line style={draw=none, insert path={(axis cs:\pgfkeysvalueof{/pgfplots/xmin},0,0) edge (axis cs:\pgfkeysvalueof{/pgfplots/xmax},0,0)}},
    	y axis line style={draw=none, insert path={(axis cs:\pgfkeysvalueof{/pgfplots/xmin},\pgfkeysvalueof{/pgfplots/ymin},0) edge (axis cs:\pgfkeysvalueof{/pgfplots/xmin},\pgfkeysvalueof{/pgfplots/ymax},0)}},
    	]
    	
    	\addplot coordinates
    	{
    		
    		(0, 2.771487)
    		(1.54, 2.849576)
    		(5.7, 2.85994)
    		(9.6, 2.800701)
    		(16.52, 2.621482)
    		(23.05, 2.447645)
    		(29.76, 2.291658)
    		(36.49, 2.141073)
    		(43.83, 1.976244)
    		(51.25, 1.809942)
    		(59.03, 1.63196)
    		(68.45, 1.431767)
    		(74.61, 1.268395)
    		(91.1, 0.579087)
    	};
    	\addplot coordinates
    	{
    		
    		(0, 2.771487)
    		(1.56, 2.894104)
    		(5.98, 2.876006)
    		(10.67, 2.791686)
    		(20.31, 2.548079)
    		(28.47, 2.357087)
    		(36.71, 2.14092)
    		(44.69, 1.927579)
    		(51.77, 1.76483)
    		(57.98, 1.640623)
    		(65.5, 1.428109)
    		(73.51, 1.255498)
    		(79.19, 1.084564)
    		(91.42, 0.552862)
    	};
    	
    	
    	\addplot[only marks,
    	mark=*,
    	mark size=3pt,
    	green,
    	nodes near coords = HRNet-32,
    	nodes near coords align={vertical},
    	point meta=y,
    	every node near coord/.append style={font=\tiny, yshift=-4.5mm, xshift=-3mm, color=black},
    	] coordinates
    	{
    		(51.57 , 1.339235)
    	};
    	
    	\addplot[
    	only marks,
    	mark=*,
    	mark size=3pt,
    	green,
    	nodes near coords = HRNet-18,
    	nodes near coords align={vertical},
    	point meta=y,
    	every node near coord/.append style={font=\tiny, yshift=-4.5mm, xshift=-3mm, color=black},
    	] coordinates
    	{
    		(80.15 , 0.829931)
    	};
    	
    	\addplot[
    	only marks,
    	mark=*,
    	mark size=3pt,
    	green,
    	nodes near coords = HRNet-48,
    	nodes near coords align={vertical},
    	point meta=y,
    	every node near coord/.append style={font=\tiny, yshift=0.5mm, xshift=5mm, color=black},
    	] coordinates
    	{
    		(0, 2.771487)
    	};
    	
    	\end{axis}
    	\begin{axis}[
    	ymin=0,
    	mark size = 0pt,
    	hide x axis,
    	axis y line*=right,
    	ylabel shift = -5 pt,
    	width=0.5\textwidth,
    	height=1.75in,
    	ylabel={mIoU (\%)},
    	ylabel near ticks,
    	dashed,
    	xtick pos=left,
    	ytick pos=right
    	]
    	
    	\addplot coordinates
    	{
    		(1.54 , 77.4)
    		(5.7 , 77.2)
    		(9.6, 77.56 ) 
    		(16.52, 76.17) 
    		(23.05, 76.10 ) 
    		(29.76, 76.78) 
    		(36.49, 76.92) 
    		(43.83, 76.29) 
    		(51.25, 76.10) 
    		(59.03, 76.17) 
    		(68.45, 74.15) 
    		(74.61 , 71.1)
    		(91.1 , 57.9)
    	};
    	\addplot coordinates
    	{
    		(1.56, 77.1)
    		(5.98, 77.1)
    		(10.67, 77.44)
    		(20.31, 77.02)
    		(28.47, 77.22)
    		(36.71, 77.35)
    		(44.69, 76.97)
    		(51.77, 76.53)
    		(57.98, 76.19)
    		(65.5, 75.77)
    		(73.51, 74.82)
    		(79.19 , 73.3)
    		(91.42 , 62.8)
    	};
    	
    	\end{axis}
    	\end{scope}
    	\end{tikzpicture}
    	\vspace{-5pt}
    	\caption{$(1\times3\times512\times1024)$}
    \end{subfigure}%
    
    \ref{named}
    
    \caption{Energy consumption, averaged over 1000 inferences, of HRNet-48, pruned at different rates. Dashed curves provide respective mIoU for each method. Green dots provide energy consumption of non-pruned network as a reference. Their position on the x-axis depends on their number of parameters (or operations) relatively to that of a non-pruned HRNet-48.}
    \label{fig:power_consumption}
\end{figure}

We measured power consumption using the tegrastats utility every second while ONNX Runtime executes 1000 inference of a given network on GPU, after 100 inferences as a warm-up. We compute the integral of instantaneous power consumption (in mW) for all the duration of the experiment and get the overall energy consumption (in J). We report the results 
for inputs of different resolutions and batch sizes in Figure~\ref{fig:power_consumption}. We chose 2 different dimensions: $(1\times3\times64\times128)$ and $(1\times3\times512\times1024)$. We also report results for non-pruned baselines such as HRNet-32 or HRNet-18.

\begin{figure}[ht]
    \centering
    \begin{adjustwidth}{-1in}{-1in}

    	\begin{subfigure}{\linewidth}
    		\centering
    		\begin{tikzpicture}
    		\begin{scope}[scale=0.9]
    		\begin{axis}[
    		ylabel=mIoU (\%),
    		mark size = 1pt,
    		ytick={60,65,70,75},
    		xlabel=Energy (J),
    		ylabel shift = -6 pt,
    		width=0.3\textwidth,
    		height=1.75in,
    		legend entries={Slimming,  SWD},
    		legend to name=named3,
    		legend columns=-1,
    		legend style = {draw=none},
    		xtick pos=left,
    		ytick pos=left
    		]
    		
    		\addplot+[mark=*] coordinates 
    		{
    			
    			(2.771487, 77.0)
    			(2.849576, 77.4)
    			(2.85994, 77.2)
    			(2.800701, 77.56)
    			(2.621482, 76.17)
    			(2.447645, 76.10)
    			(2.291658, 76.78)
    			(2.141073, 76.92)
    			(1.976244, 76.29)
    			(1.809942, 76.10)
    			(1.63196, 76.17)
    			(1.431767, 74.15)
    			(1.268395, 71.1)
    			(0.579087, 57.9)
    			
    		};
    		\addplot+[mark=*] coordinates
    		{
    			
    			(2.771487, 77.0)
    			(2.894104, 77.1)
    			(2.876006, 77.1)
    			(2.791686, 77.44)
    			(2.548079, 77.02)
    			(2.357087, 77.22)
    			(2.14092, 77.35)
    			(1.927579, 76.97)
    			(1.76483, 76.53)
    			(1.640623, 76.19)
    			(1.428109, 75.77)
    			(1.255498, 74.82)
    			(1.084564, 73.3)
    			(0.552862, 62.8)

    		};
    		
    		\pgfplotsset{cycle list shift=-2}
    		\addplot+[mark=square*] coordinates 
    		{
    			
    			(0.829931 , 74.34)
    			(0.790468, 74.86)
    			(0.70118, 74.67)
    			(0.59846, 72.38)
    			
    		};
    		
    		\addplot+[mark=square*] coordinates
    		{

    			(0.829931 , 74.34)
    			(0.783761, 75.11)
    			(0.685882, 74.31)
    			(0.584667, 72.79)
    		};
    		
    		\addplot[only marks,
    		mark=*,
    		mark size=3pt,
    		green,
    		nodes near coords = HRNet-32,
    		nodes near coords align={vertical},
    		point meta=y,
    		every node near coord/.append style={font=\tiny, yshift=0.5mm, xshift=+2mm, color=black},
    		] coordinates
    		{
    			(1.339235 , 76.27)
    		};
    		
    		\addplot[
    		only marks,
    		mark=*,
    		mark size=3pt,
    		green,
    		nodes near coords = HRNet-18,
    		nodes near coords align={vertical},
    		point meta=y,
    		every node near coord/.append style={font=\tiny, yshift=1mm, xshift=-0.5mm, color=black},
    		] coordinates
    		{
    			(0.829931 , 74.34)
    		};
    		
    		\addplot[
    		only marks,
    		mark=*,
    		mark size=3pt,
    		green,
    		nodes near coords = HRNet-48,
    		nodes near coords align={vertical},
    		point meta=y,
    		every node near coord/.append style={font=\tiny, yshift=-4.5mm, xshift=-3mm, color=black},
    		] coordinates
    		{
    			(2.771487, 77.0)
    		};

    		
    		
    		
    		
    		
    		\end{axis}
    		\end{scope}
    		\end{tikzpicture}
    		\begin{tikzpicture}
    		\begin{scope}[scale=0.9]
    		\begin{axis}[
    		ylabel=mIoU (\%),
    		mark size = 1pt,
    		xtick={0,20,40,60,80, 100},
    		ytick={60,65,70,75},
    		xlabel=Remaining operations (\%),
    		ylabel shift = -6 pt,
    		width=0.3\textwidth,
    		height=1.75in,
    		xtick pos=left,
    		ytick pos=left
    		]
    		
    		\addplot+[mark=*] coordinates
    		{
    			(100 , 77.0) 
    			(98.46, 77.4)
    			(94.3, 77.2)
    			(90.4, 77.56) 
    			(83.48, 76.17) 
    			(76.95, 76.10) 
    			(70.24, 76.78) 
    			(63.51, 76.92) 
    			(56.17, 76.29) 
    			(48.75, 76.10) 
    			(40.97, 76.17) 
    			(31.55, 74.15) 
    			(25.39, 71.1)
    			(8.9, 57.9)
    			
    		};
    		\addplot+[mark=*] coordinates
    		{
    			(100 , 77.0) 
    			(98.44, 77.1)
    			(94.02, 77.1)
    			(89.33, 77.44)
    			(79.69, 77.02)
    			(71.53, 77.22)
    			(63.29, 77.35)
    			(55.31, 76.97)
    			(48.23, 76.53)
    			(42.02, 76.19)
    			(34.5, 75.77)
    			(26.81, 74.82)
    			(20.81, 73.3)
    			(8.58, 62.8)
    			
    			(19.85 , 74.34)
    			(15.95, 75.11)
    			(12.82, 74.31)
    			(9.29, 72.79)
    		};
    		\pgfplotsset{cycle list shift=-2}
    		\addplot+[mark=square*] coordinates
    		{
    			(19.85 , 74.34)
    			(15.75, 74.86)
    			(13.06, 74.67)
    			(9.61, 72.38)
    		};
    		\addplot+[mark=square*] coordinates
    		{
    			(19.85 , 74.34)
    			(15.95, 75.11)
    			(12.82, 74.31)
    			(9.29, 72.79)
    		};
    		
    		\addplot[only marks,
    		mark=*,
    		mark size=3pt,
    		green,
    		nodes near coords = HRNet-32,
    		nodes near coords align={vertical},
    		point meta=y,
    		every node near coord/.append style={font=\tiny, yshift=-4.5mm, xshift=2mm, color=black},
    		] coordinates
    		{
    			(48.43 , 76.27) 
    			
    		};
    		
    		\addplot[
    		only marks,
    		mark=*,
    		mark size=3pt,
    		green,
    		nodes near coords = HRNet-18,
    		nodes near coords align={vertical},
    		point meta=y,
    		every node near coord/.append style={font=\tiny, yshift=1mm, xshift=-1mm, color=black},
    		] coordinates
    		{
    			(19.85 , 74.34) 
    			
    		};
    		
    		\addplot[
    		only marks,
    		mark=*,
    		mark size=3pt,
    		green,
    		nodes near coords = HRNet-48,
    		nodes near coords align={vertical},
    		point meta=y,
    		every node near coord/.append style={font=\tiny, yshift=-4.5mm, xshift=-3mm, color=black},
    		] coordinates
    		{
    			(100 , 77) 
    			
    		};
    		
    		\end{axis}
    		\end{scope}
    		\end{tikzpicture}
    		\begin{tikzpicture}
    		\begin{scope}[scale=0.9]
    		\begin{axis}[
    		ylabel=mIoU (\%),
    		mark size = 1pt,
    		xtick={0,20,40,60,80, 100},
    		ytick={60,65,70,75},
    		xlabel=Remaining parameters (\%),
    		ylabel shift = -6 pt,
    		width=0.3\textwidth,
    		height=1.75in,
    		xtick pos=left,
    		ytick pos=left
    		]
    		
    		\addplot+[mark=*] coordinates
    		{
    			(100 , 77.0) 
    			(99, 77.4)
    			(95, 77.2)
    			(90, 77.56) 
    			(80, 76.17) 
    			(70, 76.10) 
    			(60, 76.78) 
    			(50, 76.92) 
    			(40, 76.29) 
    			(30, 76.10) 
    			(20, 76.17) 
    			(10, 74.15) 
    			(5, 71.1)
    			(1, 57.9)
    			
    			
    		};
    		\addplot+[mark=*] coordinates
    		{
    			(100 , 77.0) 
    			(99, 77.1)
    			(95, 77.1)
    			(90, 77.44)
    			(80, 77.02)
    			(70, 77.22)
    			(60, 77.35)
    			(50, 76.97)
    			(40, 76.53)
    			(30, 76.19)
    			(20, 75.77)
    			(10, 74.82)
    			(5, 73.3)
    			(1, 62.8)
    			
    		};
    		\pgfplotsset{cycle list shift=-2}
    		\addplot+[mark=square*] coordinates
    		{
    			(14.64 , 74.34)
    			(10.98, 74.86)
    			(7.32, 74.67)
    			(3.66, 72.38)
    			
    		};
    		\addplot+[mark=square*] coordinates
    		{
    			(14.64 , 74.34)
    			(10.98, 75.11)
    			(7.32, 74.31)
    			(3.66, 72.79)
    		};
    		
    		\addplot[only marks,
    		mark=*,
    		mark size=3pt,
    		green,
    		nodes near coords = HRNet-32,
    		nodes near coords align={vertical},
    		point meta=y,
    		every node near coord/.append style={font=\tiny, yshift=0mm, xshift=-5mm, color=black},
    		] coordinates
    		{
    			(44.86 , 76.27) 
    			
    		};
    		
    		\addplot[
    		only marks,
    		mark=*,
    		mark size=3pt,
    		green,
    		nodes near coords = HRNet-18,
    		nodes near coords align={vertical},
    		point meta=y,
    		every node near coord/.append style={font=\tiny, yshift=-4.5mm, xshift=4mm, color=black},
    		] coordinates
    		{
    			(14.64 , 74.34) 
    			
    		};
    		
    		\addplot[
    		only marks,
    		mark=*,
    		mark size=3pt,
    		green,
    		nodes near coords = HRNet-48,
    		nodes near coords align={vertical},
    		point meta=y,
    		every node near coord/.append style={font=\tiny, yshift=-4.5mm, xshift=-3mm, color=black},
    		] coordinates
    		{
    			(100 , 77) 
    			
    		};
    		
    		\end{axis}
    		\end{scope}
    		\end{tikzpicture}
    	\end{subfigure}%

    \end{adjustwidth}
    \ref{named3}
    
    \caption{mIoU accuracy of pruned HRNet-48 (filled circles) and HRNet-18 (squares) on CityScapes, for inputs of size $(1\times3\times512\times1024)$, as a function of energy (left plot), remaining operations (middle plot) and remaining parameters (right plot). Green dots provide performance of non-pruned HRNet-48, HRNet-32 and HRNet-18 as a reference. Percentages of remaining operations or parameters are indexed on the cost of a non-pruned HRNet-48, whatever the network.}
    \label{fig:tradeoff}
\end{figure}
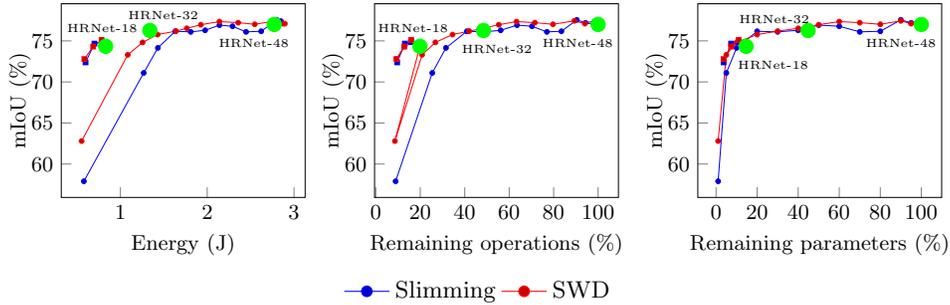

Figure~\ref{fig:tradeoff} shows various trade-offs, for inputs of size $(1\times3\times512\times1024)$,  between mIoU accuracy of the networks on CityScapes and their energy consumption, amount of remaining operations or parameters. All our reported numbers of operations do not take into account those of the \textit{ScatterND} operators, but only that of convolutions. Indeed, as energy consumption already includes the impact of these operators, our operation count rather gives an indication on the distribution of sparsity in the network.


\section{Discussion}

\subsection{Impact on Hardware}

\begin{table}[hb]
    \centering
    \begin{tabularx}{\textwidth}{XXXXXXXXX}
    	\toprule
    	&Param.& Op. & Lat. & Lat.* & C. Lat. &Lat. & Lat.* & C. Lat. \\
    	&&&\multicolumn{3}{c}{$(1\times 3\times 64 \times 128)$}&\multicolumn{3}{c}{$(1\times 3\times 512 \times 1024)$}\\
    	SWD& 40.0\% & 48.2\% & 25.1ms & 20.3ms & 19.9ms & 245.2ms & 196.0ms & 188.0ms\\
    	Slim.& 30.0\% & 48.8\% & 22.5ms & 17.9ms & 17.5ms & 248.1ms & 199.8ms & 191.9ms\\
    	HRN32 & 44.9\% & 48.4\% & 15.1ms& 15.1ms& 14.4ms & 158.3ms &  158.3ms & 147.7ms\\
    	\bottomrule
    \end{tabularx}
    \caption{Count of remaining parameters (Param.) and operations (Op.), overall latency (Lat.), latency with the exclusion of foreign nodes (Lat.*) and latency of convolution layers only (C. Lat.), for two different sizes of inputs, of HRNet-48 pruned using SWD (SWD) or Slimming (Slim.) and a non-pruned HRNet-32 (HRN32). For HRNet-32, the percentage of parameters and operations gives a relative comparison to those of non-pruned HRNet-48. Measurements and profiling were performed using the \textit{trtexec} utility.}
    \label{tab:trtexec}
\end{table}

Our experiments show that we can leverage global filter pruning and reduce energy consumption. The efficiency of pruning slightly depends on the size of the inputs, even though results get more stable once inputs are of sufficient resolution, as can be seen in Figure~\ref{fig:power_consumption}. This was achieved using \textit{ScatterND} ONNX operations, whose cost in latency and energy is not negligible. Figure~\ref{fig:power_consumption} shows that non-pruned smaller networks, such as HRNet-32 and HRNet-18, are energetically more efficient than pruned networks, even when comparing at equal number of operations. One would expect operations, latency and energetic consumption to be correlated, and that the difference between, for example, the consumption and latency of HRNet-32 and that of pruned networks of similar count of operations should match. However, this is only partly true, as indicated in Table~\ref{tab:trtexec}.

Indeed, once the cost of the ``foreign nodes'', that are what TensorRT turn the \textit{ScatterND} operations into, is removed from the overall latency of the network, results in latency are closer to each other but there is still a difference. When isolating the latency of convolution layers, that are supposed to count the same number of operations across the three compared networks, one can see that those of non-pruned HRNet-32 run significantly faster. As the difference in operations, that is minimal, or parameters, that should be in favor to the pruned networks, cannot explain this gap, we suppose that it is due to differences in the acceleration tactics empirically chosen by TensorRT. This would mean that not only \textit{ScatterND} operations are costly but also that their presence harms the optimization of the rest of the network. This is likely due to their unconventional nature in convolution networks.

The most likely solutions to this problem would be either: 1) a better choice of operators, that we do not know as our solution is the one that gave us the best performance, 2) a new custom operator that cuts the drawbacks of our current solution, that uses preexisting ONNX operators or 3) a better support of \textit{Scatter} operations and of this type of use in networks from TensorRT and other frameworks.

\subsection{Pruning as Architecture Search}\label{archi}

Figure~\ref{fig:tradeoff} provides three different trade-off curves, for both pruned HRNet-48 and HRNet-18, which each lead to a different observation. Green dots provide performance of non-pruned networks. The accuracy-to-energy trade-off as well as the accuracy-to-operations trade-off is in favor of HRNet-18 compared to HRNet-48. On the contrary, HRNet-48 is better on the accuracy-to-parameters trade-off. In the literature on network pruning, the parameters count is often used to measure the complexity of the pruned network. Our results show that it is a bad proxy for its actual energy consumption and computational complexity. The reason is that, since different layers in the network are applied to intermediate representations of different resolutions, all parameters in the network do not have the same cost in term of operations and energy.

It seems that pruning HRNet-48 leads to save parameters that are responsible for many operations. Indeed, HRNet-18, that is uniformly thinner than HRNet-48, counts much fewer operations than pruned networks that contain as many parameters. This is surprising, since the criterion described in Section~\ref{strategies} does not explicitly imply any bias towards targeting parameters responsible for fewer or more operations. Since this behavior leads to architectures that seem sub-optimal, one can wonder if this the best that pruning can do or if it is due to some unexpected bias in the criterion.

One hypothesis would be that parameters that are responsible for many operations are indeed more important and of greater magnitude, in which case the count of operations and accuracy would be strongly correlated. This hypothesis is rather unlikely. Indeed, one example of strong correlation is the parameters-to-accuracy trade-off in Figure~\ref{fig:tradeoff}: all points are roughly located on the same average curve, whatever the pruning method or the pruned network. In the case of the operations-to-accuracy (or even the energy-to-accuracy) trade-off, there is a noticeable gap between pruned HRNet-48 and HRNet-18 as well as between networks pruned using SWD or Slimming. Moreover, networks pruned using SWD both contain fewer operations and have a better accuracy compared to those pruned using Slimming, even though they contain the same number of parameters, which is the opposite behavior to the one to be expected if the hypothesis was true.

If such parameters are not more important, then there is apparently no reason why pruning would prioritize saving them. Therefore it is likely that such a behavior is an accident and due to an unwanted bias in the criterion. Various types of analogous imbalances are to be seen in many criteria in the literature~\cite{tanaka2020pruning}. Since this bias visibly harms pruning in a way that makes it less efficient than training thin baselines from scratch, it seems that further investigations are necessary to figure out the reason of such biases and how to avoid them to improve considerably the performance of pruning.

Finally, the fact that pruning HRNet-18 leads to better results than pruning HRNet-48 leads to both the intuitive observation, that it is better to prune a network that is closer to the desired target, and the counter-intuitive observation that pruning the bigger network failed to come up with the same efficient architecture, while the ability of pruning to produce them is supposedly its main interest~\cite{liu2018rethinking}.


\section{Conclusion}

We have measured the energetic consumption of the HRNet-48 and HRNet-18 semantic segmentation networks, pruned using global structured pruning methods. Thanks to the use of \textit{ScatterND} ONNX operations, we are able to leverage the sparsity on embedded GPU, demonstrating the ability of global pruning to reduce the cost of networks. However, our experiments also show the necessity both to elaborate a more efficient implementation of operators required to leverage pruning and to explore further how biased are pruning criteria, as they lead to distributions of sparsity that are energetically sub-optimal.

\bibliographystyle{plain}
\bibliography{SysInt_paper}

\begin{thebibliography}{10}

\bibitem{badrinarayanan2017segnet}
Vijay Badrinarayanan, Alex Kendall, and Roberto Cipolla.
\newblock Segnet: A deep convolutional encoder-decoder architecture for image
  segmentation.
\newblock {\em IEEE transactions on pattern analysis and machine intelligence},
  39(12):2481--2495, 2017.

\bibitem{chen2017deeplab}
Liang-Chieh Chen, George Papandreou, Iasonas Kokkinos, Kevin Murphy, and Alan~L
  Yuille.
\newblock Deeplab: Semantic image segmentation with deep convolutional nets,
  atrous convolution, and fully connected crfs.
\newblock {\em IEEE transactions on pattern analysis and machine intelligence},
  40(4):834--848, 2017.

\bibitem{cordts2016cityscapes}
Marius Cordts, Mohamed Omran, Sebastian Ramos, Timo Rehfeld, Markus Enzweiler,
  Rodrigo Benenson, Uwe Franke, Stefan Roth, and Bernt Schiele.
\newblock The cityscapes dataset for semantic urban scene understanding.
\newblock In {\em Proceedings of the IEEE conference on computer vision and
  pattern recognition}, pages 3213--3223, 2016.

\bibitem{courbariaux2015binaryconnect}
Matthieu Courbariaux, Yoshua Bengio, and Jean-Pierre David.
\newblock Binaryconnect: Training deep neural networks with binary weights
  during propagations.
\newblock {\em Advances in neural information processing systems}, 28, 2015.

\bibitem{han2015deep}
Song Han, Huizi Mao, and William~J Dally.
\newblock Deep compression: Compressing deep neural networks with pruning,
  trained quantization and huffman coding.
\newblock {\em arXiv preprint arXiv:1510.00149}, 2015.

\bibitem{han2015learning}
Song Han, Jeff Pool, John Tran, and William Dally.
\newblock Learning both weights and connections for efficient neural network.
\newblock {\em Advances in neural information processing systems}, 28, 2015.

\bibitem{he2016deep}
Kaiming He, Xiangyu Zhang, Shaoqing Ren, and Jian Sun.
\newblock Deep residual learning for image recognition.
\newblock In {\em Proceedings of the IEEE conference on computer vision and
  pattern recognition}, pages 770--778, 2016.

\bibitem{he2017channel}
Yihui He, Xiangyu Zhang, and Jian Sun.
\newblock Channel pruning for accelerating very deep neural networks.
\newblock In {\em Proceedings of the IEEE international conference on computer
  vision}, pages 1389--1397, 2017.

\bibitem{hinton2015distilling}
Geoffrey Hinton, Oriol Vinyals, Jeff Dean, et~al.
\newblock Distilling the knowledge in a neural network.
\newblock {\em arXiv preprint arXiv:1503.02531}, 2(7), 2015.

\bibitem{hong2021deep}
Yuanduo Hong, Huihui Pan, Weichao Sun, and Yisong Jia.
\newblock Deep dual-resolution networks for real-time and accurate semantic
  segmentation of road scenes.
\newblock {\em arXiv preprint arXiv:2101.06085}, 2021.

\bibitem{lecun1989optimal}
Yann LeCun, John Denker, and Sara Solla.
\newblock Optimal brain damage.
\newblock {\em Advances in neural information processing systems}, 2, 1989.

\bibitem{liu2021s2ta}
Zhi-Gang Liu, Paul~N Whatmough, Yuhao Zhu, and Matthew Mattina.
\newblock S2ta: Exploiting structured sparsity for energy-efficient mobile cnn
  acceleration.
\newblock {\em arXiv preprint arXiv:2107.07983}, 2021.

\bibitem{liu2017learning}
Zhuang Liu, Jianguo Li, Zhiqiang Shen, Gao Huang, Shoumeng Yan, and Changshui
  Zhang.
\newblock Learning efficient convolutional networks through network slimming.
\newblock In {\em Proceedings of the IEEE international conference on computer
  vision}, pages 2736--2744, 2017.

\bibitem{liu2018rethinking}
Zhuang Liu, Mingjie Sun, Tinghui Zhou, Gao Huang, and Trevor Darrell.
\newblock Rethinking the value of network pruning.
\newblock {\em arXiv preprint arXiv:1810.05270}, 2018.

\bibitem{long2015fully}
Jonathan Long, Evan Shelhamer, and Trevor Darrell.
\newblock Fully convolutional networks for semantic segmentation.
\newblock In {\em Proceedings of the IEEE conference on computer vision and
  pattern recognition}, pages 3431--3440, 2015.

\bibitem{ma2021non}
Xiaolong Ma, Sheng Lin, Shaokai Ye, Zhezhi He, Linfeng Zhang, Geng Yuan,
  Sia~Huat Tan, Zhengang Li, Deliang Fan, Xuehai Qian, et~al.
\newblock Non-structured dnn weight pruning--is it beneficial in any platform?
\newblock {\em IEEE Transactions on Neural Networks and Learning Systems},
  2021.

\bibitem{renda2020comparing}
Alex Renda, Jonathan Frankle, and Michael Carbin.
\newblock Comparing rewinding and fine-tuning in neural network pruning.
\newblock {\em arXiv preprint arXiv:2003.02389}, 2020.

\bibitem{ronneberger2015u}
Olaf Ronneberger, Philipp Fischer, and Thomas Brox.
\newblock U-net: Convolutional networks for biomedical image segmentation.
\newblock In {\em International Conference on Medical image computing and
  computer-assisted intervention}, pages 234--241. Springer, 2015.

\bibitem{simonyan2014very}
Karen Simonyan and Andrew Zisserman.
\newblock Very deep convolutional networks for large-scale image recognition.
\newblock {\em arXiv preprint arXiv:1409.1556}, 2014.

\bibitem{sun2019high}
Ke~Sun, Yang Zhao, Borui Jiang, Tianheng Cheng, Bin Xiao, Dong Liu, Yadong Mu,
  Xinggang Wang, Wenyu Liu, and Jingdong Wang.
\newblock High-resolution representations for labeling pixels and regions.
\newblock {\em arXiv preprint arXiv:1904.04514}, 2019.

\bibitem{tanaka2020pruning}
Hidenori Tanaka, Daniel Kunin, Daniel~L Yamins, and Surya Ganguli.
\newblock Pruning neural networks without any data by iteratively conserving
  synaptic flow.
\newblock {\em Advances in Neural Information Processing Systems},
  33:6377--6389, 2020.

\bibitem{tessier2022rethinking}
Hugo Tessier, Vincent Gripon, Mathieu L{\'e}onardon, Matthieu Arzel, Thomas
  Hannagan, and David Bertrand.
\newblock Rethinking weight decay for efficient neural network pruning.
\newblock {\em Journal of Imaging}, 8(3):64, 2022.

\bibitem{yu2015multi}
Fisher Yu and Vladlen Koltun.
\newblock Multi-scale context aggregation by dilated convolutions.
\newblock {\em arXiv preprint arXiv:1511.07122}, 2015.

\bibitem{yuan2020object}
Yuhui Yuan, Xilin Chen, and Jingdong Wang.
\newblock Object-contextual representations for semantic segmentation.
\newblock In {\em European conference on computer vision}, pages 173--190.
  Springer, 2020.

\bibitem{zhao2019region}
Shuai Zhao, Yang Wang, Zheng Yang, and Deng Cai.
\newblock Region mutual information loss for semantic segmentation.
\newblock {\em Advances in Neural Information Processing Systems}, 32, 2019.

\end{thebibliography}
\end{document}